\documentclass{article}

\usepackage{arxiv}
\usepackage[utf8]{inputenc} 
\usepackage[T1]{fontenc}    
\usepackage{hyperref}       
\usepackage{url}            
\usepackage{booktabs}       
\usepackage{amsfonts}       
\usepackage{nicefrac}       
\usepackage{microtype}      
\usepackage{lipsum}
\usepackage{graphicx}
\usepackage{float}
\usepackage{array}
\usepackage{fancyhdr}       
\graphicspath{ {./images/} }

\title{Predicting Emotion Intensity in Polish Political Texts: Comparing Supervised Models and Large Language Models in a Resource-Poor Language}

\author{
  Hubert Plisiecki \\
  Institute of Psychology\\
  Polish Academy of Sciences\\
  \texttt{hplisiecki@gmail.com} \\
  ORCID: 0000-0002-5273-1716 \\
  \And
  Piotr Koc \\
  GESIS – Leibniz Institute for the Social Sciences\\
  \texttt{pkoc@gesis.org} \\
  ORCID: 0000-0003-2152-4594 \\
  \And
  Maria Flakus \\
  Institute of Philosophy and Sociology\\
  Polish Academy of Sciences\\
  \texttt{mflakus@ifispan.edu.pl} \\
  ORCID: 0000-0002-6667-8020 \\
  \And
  Artur Pokropek \\
  Institute of Philosophy and Sociology\\
  Polish Academy of Sciences\\
  \texttt{apokropek@ifispan.edu.pl} \\
  ORCID: 0000-0002-5899-2917 \\
}

\begin{document}
\maketitle
\begin{abstract}
This study explores the use of large language models (LLMs) to predict emotion intensity in Polish political texts, a resource-poor language context. The research compares the performance of several LLMs against a supervised model trained on an annotated corpus of 10,000 social media texts, evaluated for the intensity of emotions by expert judges. The findings indicate that while the supervised model generally outperforms LLMs, offering higher accuracy and lower variance, LLMs present a viable alternative, especially given the high costs associated with data annotation. The study highlights the potential of LLMs in low-resource language settings and underscores the need for further research on emotion intensity prediction and its application across different languages and continuous features. The implications suggest a nuanced decision-making process to choose the right approach to emotion prediction for researchers and practitioners based on resource availability and the specific requirements of their tasks.
\end{abstract}


\section{Introduction}
\subsection{Significance of emotions in psychological science and present advancements in research on emotions}
Over the past few decades, social scientists have broadened their research to understand the significant role that emotions play in human behavior and societal dynamics. This exploration has yielded important findings in political sciences (Mintz et al., 2022), sociology (Bericat, 2016; Turner \& Stets, 2006), economics (Loewenstein, 2000), anthropology (Lutz \& White, 1986), organizational research (Diener et al., 2020) as well other fields of social (Kleef, 2018) and psychological sciences (Derks et al., 2008).

While investigating this role of emotions, many researchers concentrate on the question of whether an emotion is present, focusing on the categorical aspects of emotions (Fritz et al., 2009; Saarimäki et al., 2016; Siedlecka \& Denson, 2019; Tanaka-Matsumi et al., 1995). However, beyond the sole presence or absence of emotions, there is also their intensity,  which was early recognized as necessary to understand human behaviors (Brehm, 1999; Plutchik, 1965). People often describe emotions like anger, sadness, or happiness in varying degrees, from none at all to very intense, and research indicates that emotion intensity is crucial in cognitive processing, social behavior, and communication within groups (Frijda et al., 1992; Niedenthal \& Brauer, 2012; Reisenzein, 1994). 

There is also the third general approach to studying emotions, i.e., dimensional models. In contrast to categorical and intensity approaches, dimensional models offer a different perspective by suggesting that emotions can be placed within a continuous space defined by specific dimensions representing fundamental properties of emotional states (Gendron \& Feldman Barrett, 2009). The most recognized model from this branch is Russell's (1980) circumplex model of affect, which posits that all emotions can be characterized by two fundamental dimensions: valence, which is the degree of pleasure or displeasure, and arousal, the level of activation or deactivation associated with an emotion. 

Access to numerous text data sources, including social media content, responses to open-ended questions in computer-based assessments, political declarations, newspapers, and online forums, offers an unprecedented opportunity to study emotions beyond the traditional settings of psychological laboratories. To capture emotions in text, scholars initially focused on valence, i.e., differentiating between positive and negative sentiment. However, an increasing body of research has demonstrated that emotions of the same valence can affect social processes in different ways (Druckman \& McDermott, 2008; Nabi, 2003; Valentino et al., 2011; Vasilopoulos et al., 2019) and that those distinct (discrete) emotions, like anger and happiness, can be identified in text (e.g., Pennebaker \& Francis, 1996).  As a result, various tools for discrete emotion detection were created, mainly for the English language context, with far fewer tools available for other languages (Mohammad, 2016; Üveges \& Ring, 2023). The emotion intensity approach has been largely overlooked in natural language processing (NLP) applications. While some attempts at predicting intensity exist, they are very rare (e.g. Akhtar et al., 2020). We can attribute it to the straightforwardness of the discrete approach. For instance, annotating items regarding the binary occurrence of emotions is way easier than their intensity. 

Recently, large language models (LLMs) have contributed to advancements in NLP, including emotion classification. These models have demonstrated their effectiveness in accurately identifying discrete emotions in the text by leveraging vast data and complex pattern recognition capabilities (Kocoń et al., 2023). Their success in this domain suggests the potential of LLMs to also aid in predicting emotion intensity, which has not yet been thoroughly investigated. The approach using LLMs is an auspicious direction for “resource-poor languages” (Mohammad 2016, 203), where researchers often encounter the problem of lacking adequate tools for analyzing emotions. A problem, which is difficult to solve as the expression of emotions is language (Bazzanella, 2004) and domain (Haselmayer \& Jenny, 2017; Rauh, 2018) specific, which requires researchers to use or create linguistically adapted tools for a particular kind of corpora (Üveges and Ring 2023). 

In this work, we explore the potential of LLMs to replace human annotators and traditional predictive models in the task of predicting emotion intensity in one of the resource-poor languages, Polish, focusing on political texts. To do so, we build a corpus of political texts using different social media sources and have 10, 000 texts annotated by expert judges, whose reliability in assessing the intensity of emotions is evaluated. Then, we compare the performance of several LLMs to a supervised model that was trained on annotated datasets in predicting the intensity of emotions. 

The results show that the supervised model trained on the annotated data generally outperforms the LLMs, offering marginally higher accuracy and lower variance. However, this comes at the cost of the resources needed for the data annotation. Overall, the findings hold promise for using LLMs to assess other continuous features in Polish and potentially extend to other resource-poor languages.
\subsection{Emotions and social media - previous research}
Researching emotions in social media is of high importance as these social platforms have evolved into significant channels for spreading opinions and emotions (Beskow \& Carley, 2019) - primarily due to their striking popularity, with 77.8\% of people over the age of 18 using them  globally (DataReportal, 2023). Research indicates that social media emotions may motivate people to share certain content (Brady et al., 2017), buy commercial products (Lu et al., 2021), change their behaviors (McBride \& Ball, 2022), or even divide and disrupt populations (Whitehead, 2016).

It also has been demonstrated that a reaction to social media posts may depend on the specific emotions enhanced by this content and the context of its origin. As for the former, for instance, emotions rich in arousal (e.g., anger and anxiety) may increase information sharing, regardless of emotional valence (Berger \& Milkman, 2012; Stieglitz \& Dang-Xuan, 2013). As for the latter, for example, in political discourse, content sharing is more probable if it includes fear bait or support communication (Walker et al., 2017). 

However, although there is general scientific consensus regarding the great utility of emotion evaluation in social media texts, the studies that have done so to date have been largely limited. They have either (1) measured emotion to a limited extent (e.g., limiting the number of estimated emotions or keywords to refer to them) or (2) failed to capture the nuances of emotional responses (e.g., their dimensional nature and intensity), ignoring emotions' complexity (Elfenbein \& Ambady, 2002). 

Regarding the first point, current research acknowledges that affective and emotional responses may capture basic emotions ("discrete emotions"; e.g., happiness, sadness, and anger), as well as their combinations, often discussed as "moods" (Ekman, 1992a), and are universal, culturally- and situation-dependent (Russell, 2014). Nevertheless, there is still open discussion regarding the exact number of emotions that should be distinguished, starting from Ekman's (1992a,b) theory of six basic emotions (enlisting happiness, anger, sadness, fear, disgust, and surprise), and ending with his notable critiques (Barrett et al., 2019; Elfenbein \& Ambady, 2002). For instance, Barrett et al. (2019) opted for more emotional dimensions (over 20). Correspondingly, other researchers opted for describing emotions using more specific categories (e.g., irritation or rage) rather than its more extensive "umbrella terms", e.g., anger (Cowen \& Keltner, 2017), especially because some specific emotions, for instance, awe and wonder (Keltner \& Haidt, 2003; Shiota et al., 2007), compassion, sympathy, and empathic pain (Goetz et al., 2010) may not fully comprehend with generalized categories. As there is no clear consensus in this discussion, the researcher's choice regarding the number of emotions may also be seen as a reflection of their theoretical or ontological preferences.

Regarding the latter, second point, ignoring emotions' complexity and dimensionality may potentially harm the validity of conclusions, as it "tears down" the measurement of the theoretical nature of emotional responses - contrary to the choice regarding the number of emotions, which may only limit the interpretation of the results to the exact dimensions which were under investigation in a particular study (Paletz et al., 2023). Indeed, some of the previously described annotation schemes do not measure emotions' intensity (Alm et al., 2005; Novielli et al., 2018) or operate on exclusive coding, in which only one emotion may be chosen per text (Alm et al., 2005).

In the current research we try to capture the emotional content of social media posts in a more robust way, although the list of emotions that we try to map is limited, we believe that it is comprehensive and additionally we fully acknowledge the arbitrary choice of the specific emotions that we have picked. We combine the dimensional, circumplex model with basic emotions, and additionally predict the intensity of the latter, which so far has been rarely attempted.
\subsection{Emotions intensity - promising research gap or scientific dead end?}
The intensity of emotion was recognized early as an important and natural extension of the basic classification scheme in both theoretical and practical contexts (Ferrara \& Yang, 2015; Qiu et al., 2020). One of the earlier significant attempts to use continuous emotion metrics was made by Strapparava and Mihalcea (2007). However, their results were not entirely valid. The dataset they used consisted of news headlines from major outlets like the New York Times, CNN, and BBC News, as well as from the Google News search engine. They prepared two datasets: a development dataset with 250 annotated headlines and a test dataset with 1,000 annotated headlines. Annotators were provided with six predefined emotion labels (i.e., anger, disgust, fear, joy, sadness, surprise) and asked to classify the headlines with the appropriate emotion label and/or with a valence indication (positive/negative). Additionally, an intensity scale ranging from 0 to 100 was added. The agreement between the six annotators on the emotions (calculated as the Pearson’s correlation of their scores to the averaged scores of the other annotators) was as follows: 0.50 for anger, 0.44 for disgust, 0.64 for fear, 0.60 for joy, 0.68 for sadness, 0.36 for surprise, and 0.78 for valence.

Using NLP and a lexicon-based method, they were able to detect emotions with high accuracy (binary classification): 93.6\% for anger, 97.3\% for disgust, 87.9\% for fear, 82.2\% for joy, 89.0\% for sadness, and 89.1\% for surprise. However, the performance of automated systems for emotion intensity prediction, calculated as the correlation between the original scores and the system predictions, was low: 0.32 for anger, 0.19 for disgust, 0.45 for fear, 0.26 for joy, 0.41 for sadness, and 0.17 for surprise. 

The next significant study on emotion intensity was conducted by Mohammad and Bravo-Marquez in (2017b). They created the first dataset called the “Tweet Emotion Intensity Dataset”, which consisted of 7,097 tweets annotated regarding anger, fear, joy, and sadness intensities. The reliability of annotation for intensity, supported by best-worst scaling (BWS) technique, showed high Pearson correlation coefficients: 0.80 for anger, 0.85 for fear, 0.88 for joy, and 0.85 for sadness.

In the shared task using this dataset, 22 teams participated (Mohammad \& Bravo-Marquez, 2017a), with the best-performing system achieving a Pearson correlation of 0.747 with the gold intensity scores, indicating that predicting emotion intensity is possible but challenging. Akhtar et al. (2020b) achieved the following correlations for predicted emotions with annotated emotions: 0.75 for anger, 0.71 for joy, 0.76 for sadness, and 0.78 for fear, with an average correlation of 0.75. This demonstrates that predicting the intensity of emotions, although difficult, is feasible with a reasonable level of reliability. However, this task is significantly more challenging than simple classification.

Despite the apparent success in predicting emotion intensity, further research in this area has been limited, with few exceptions where emotion intensity has been applied to the study of empirical problems (Sharifirad et al., 2019). This leaves the intensity of emotions as a theoretically valid yet rarely explored area in emotion sentiment analysis.

\subsection{LLMs as a method of classifying emotions - previous research}
LLM’s have been successfully used to predict some dimensions of emotions in text snippets. One of the experiments with Open AI models, both GPT3.5 and GPT4, have tested a variety of different annotation tasks, sentiment analysis and emotion recognition included and showed promising results, however its performance did not match that of the available State of the Art (SOTA) models at the time (Kocoń et al., 2023), and compared to other annotation tasks fared poorly especially on those tasks which were related to emotion annotation where the difference between its results and those of the SOTA models ranged from 71.3\% to 21.8\%. This result has been confirmed by other research projects, where the models developed by Open AI have also fallen short of the SOTA (Amin et al., 2023; Krugmann \& Hartmann, 2024). This however should not be interpreted as a rule for all LLMs as Amin and his team (2023) showed that the Llama model developed by Meta can match the SOTA performance on some benchmarks. 

While the models developed by Open AI might not provide the best results with regards to English benchmarks, they have been shown to be superior for the task of cross-lingual sentiment analysis (Přibáň et al., 2024) owing largely to their vast multilingual training data. For example, while the Llama model has been shown to be superior for some English emotion related tasks, due to its limited training set compared to the OpenAI models it fared worse on multilingual tasks.  For that reason in the current study, we choose to focus on the performance of GPT3.5 and GPT4 models. Furthermore, the bar set by the SOTA models for the utilization of LLMs in resource-poor languages is considerably lower as the lack of the resources also leads to lower SOTA performance.

As LLMs can accept context alongside the task that they are supposed to complete, the In-Context Learning (ICL) technique has been used repeatedly to enhance their performance (Chochlakis et al., 2024; Kocoń et al., 2023). This method relies on providing examples of the items the LLM is supposed to annotate, alongside their ground truth values in order to guide the model towards better solutions. It is also often referred to as multi-shot prediction. While this technique indeed has elevated the accuracy of LLM predictions for the most part, deeper analysis has shown that the model in many cases does not learn from the provided ground truth, but rather pays attention to the examples alone, which in turn prime the model towards similar examples that it has learned from its training set, resulting in better performance (Chochlakis et al., 2024). This could mean that multi-shot prompting should be less performant for low-resource languages as the model has been trained on comparatively less texts associated with them. While testing this hypothesis directly is beyond the scope of this research as we lack reliable control groups, we do employ multi-shot prompting in order to push the LLM to the edge of its performance, whether it works.

\section{Materials and Methods}
\subsection{Database preparation}
Our research utilizes a comprehensive database of Polish political texts from social media profiles (i.e., YouTube, Twitter, Facebook) of 25 journalists, 25 politicians, and 19 non-governmental organizations (NGOs). The complete list of the profiles is available in the Appendix. For each profile, all available posts from each platform were scraped (going back to the beginning of 2019). In addition, we also used corpora, which consists of texts written by “typical” social media users, i.e., non-professional commentators of social affairs. Our data consists of 1,246,337 text snippets (Twitter: 789490 tweets; Youtube: 42252 comments; Facebook: 414,595 posts). 

As transformer models have certain limits, i.e., their use imposes limits on length, we implemented two types of modification within the initial dataset. First, since texts retrieved from Facebook were longer than the others, we have split them into sentences. Second, we deleted all texts that were longer than 280 characters. 

The texts were further cleaned from social media artifacts, such as dates scrapped alongside the texts. Next, the langdetect (Danilak, 2021) software was used to filter out text snippets that were not written in Polish. Also, all online links and user names in the texts were replaced with “\_link\_” and  “\_user\_”, respectively, so that the model does not overfit the sources of information nor specific social media users. 

Because most texts in the initial dataset were emotionally neutral, we filtered out the neutral texts and included only these snippets which had higher emotional content in the final dataset. Accordingly, the texts were stemmed and subjected to a lexicon analysis (Imbir, 2016) using lexical norms for valence, arousal, and dominance - the three basic components of emotions. The words in each text were summed up in terms of their emotional content extracted from the lexical database and averaged to create separate metrics for the three emotional dimensions. These metrics were then summed up and used as weights to choose 8,000 texts for the final training dataset. Additionally, 2,000 texts were selected without weights to ensure the resulting model could process both neutral and emotional texts. The proportions of the texts coming from different social media platforms reflected the initial proportions of these texts, resulting in 496 YouTube texts, 6,105 Twitter texts, and 3,399 Facebook texts. 

\subsection{Annotation Process}
The final dataset consisting of 10,000 texts was annotated by 20 expert annotators (age: M = 23.89, SD = 4.10; gender: 80\% female). All annotators were well-versed in Polish political discourse and were students of Psychology (70\% of them were graduate students, which in the case of Polish academic education denotes people studying 4th and 5th year). Thus, they underwent at least elementary training in psychology.

The entire annotation process lasted five weeks. Each week, every annotator was given five sets of texts (out of 100 sets with 100 randomly assigned sentences each) that should be annotated in the given week. The sets were randomly assigned to annotators, considering the general assumption that five different annotators should annotate each set. Generally, annotators simultaneously annotated no more than 500 texts each week, preventing them from cognitive depletion's negative effects. 

Annotators labeled each text based on the five basic emotions: happiness, sadness, anger, disgust, and fear. In addition, annotators were asked to label the texts with regard to an additional emotion, namely pride, and two general dimensions of emotions: valence and arousal. In all cases, annotators used a 5-point scale (in the case of emotions: 0 = emotion is absent, 4 = very high level of emotion; in the case of valence and arousal, we used a pictographic 5-point scale provided in the Appendix).

Since two additional emotional dimensions might not have been familiar to annotators, before the formal annotation process began, all annotators were informed about the characteristics of valence and arousal (note that we did not provide formal definitions of basic emotions). General annotation guidelines were provided to ensure consistency and minimize subjectivity (all instructions used within the training process are available in the Appendix).
\subsection{Statistical analyses}
\subsubsection{Annotation Agreement}
We assessed the agreement between raters using the intraclass correlation coefficient (ICC). The ICC coefficients are based on the random model ANOVA for independent groups (McGraw \& Wong, 1996; Shrout \& Fleiss, 1979). ICC(1) measures the reliability of single-ratings. ICC(1) compares the variability between raters to the total variability across all ratings. It assesses how much of the total variance in the scores is due to the variance between the rated texts. It assumes that a different rater rates each text, and the raters are randomly selected. It determines the consistency of raters' evaluations across texts when a randomly selected rater assesses each text. The ICC(1,k) coefficient extends the concept of single-rating reliability, as measured by ICC(1), to scenarios where the average ratings from a set of k raters evaluate each subject. Specifically, it assesses the absolute agreement among these raters, considering the mean of their ratings for each text. This approach acknowledges the increased reliability expected when aggregating evaluations from multiple raters. The ICC values range from 0 to 1, with 0 indicating no agreement among raters and 1 indicating perfect reliability. Koo and Li (2016) provide a guideline for interpreting ICC values, categorizing them as follows: values below 0.50 are considered poor; values ranging from 0.50 to 0.75 indicate moderate reliability; values between 0.75 and 0.90 suggest good reliability; and values above 0.90 are deemed excellent. To estimate the ICC, we used the pingouin Python package (Vallat, 2018).
\subsubsection{Data for training, validation and testing}
After the annotation steps, we averaged the annotations corresponding to specific emotional metrics for each text. As the emotional load of the texts was still highly skewed towards lack of emotions, z scores for all of the emotions were computed, summed up, and used as weights to sample the test set, which constituted 10\% of the total dataset. We did this to prevent the model from overfitting the lack of emotions by assigning low emotions to every predicted text. The remaining data was split into a training set and validation set, rearing a split of (8:1:1). 
\subsubsection{Model Architecture}
We considered two alternative base models: the Trelbert transformer model developed by a team at DeepSense (Szmyd et al., 2023), and the Polish Roberta model (Dadas, 2020). The encoders of both models were each equipped with an additional regression layer with a sigmoid activation function. The maximum number of epochs in each training run was set to 100. At each step, we computed the mean correlation of the predicted metrics with their actual values on the evaluation batch, and the models with the highest correlations on the evaluation batch were saved to avoid overfitting. We used the MSE criterion to compute the loss alongside the AdamW optimizer with default hyperparameter values. Both of the base models were then subjected to a Bayesian grid search using the WandB platform (Wandb/Wandb, 2017/2024) with the following values:  dropout - 0; 0.20, 0.40, 0.60; learning rate - 5e-3, 5e-4, 5e-5; weight decay - 0.10, 0.30, 0.50; warmup steps - 300, 600, 900. The model which obtained the highest correlation relied on the Roberta transformer model and had the following hyperparameters: dropout = 0.6; learning rate = 5e-5; weight decay = 0.3.
\subsubsection{Robustness Analysis}
To assess the robustness of the model when trained on different subsets of the data, we performed a k-fold analysis with the same parameters as those chosen through the Bayesian grid search. We split the dataset into ten folds. On each iteration, one partition was held out, and the rest were split into the training and validation set (889 to 111 ratio to ensure approximately the exact size of the validation and test set). Then, we trained the model using the exact same method as described in the Model Architecture section.
\subsubsection{LLM Testing}
To assess the ability of LLMs to annotate the dataset properly, we have queried both gpt3\_5\_turbo\_0125 (GPT3.5) and gpt-4-0613 (GPT4) with the multiple shot technique. Also, we have tested the GPT3.5 on the zero, one, and up to five-shot setup to estimate the best-performing multiple-shot setup. The tests have been completed on the validation set in order not to overfit the test set. The discrete emotions were tested with the following query (The prompts have been translated for the purpose of presentation):

Translation:

\textit{"To what extent does the text below manifest the emotion '{emotion}'? Respond using a 5-point scale, where 1 means the emotion is not present at all and 5 means the emotion is very distinctly present. Please respond with a single number. Text: '{text}' Your response:"}

While the dimensions of valence and arousal had these prompts:

Valence:

\textit{"What emotional valence do you read in the following text? Respond using a 5-point scale, where 1 indicates a negative emotion is present and 5 indicates a positive emotion is present. Please respond with a single number."}

Arousal:

\textit{"What level of arousal do you read in the following text? Respond using a 5-point scale, where 1 means no arousal and 5 means extreme arousal. Please respond with a single number."}

Due to the difference in the prompts as well as the qualitative difference between the dimensions and basic emotion, we have conducted two separate tests for each type of emotion taxonomy (basic vs dimensional affective metrics). The prompts were created based on the questions that annotators provided during the annotation process. They were structured in accordance with the official OPENAI prompt engineering guidelines (OpenAI Platform, n.d.). For an in-depth description of how the prompts were structured see Appendix.

The examples for the multiple-shot scenarios were picked in the following manner. First, we have vectorized the training set using the text-embedding-3-small model from the OPENAI API. Based on the resulting vectors, we calculated the centroid of the embeddings to represent the central point of our dataset. We then determined each text's distance from this centroid to assess its representativeness or deviation from the rest of the texts in the dataset. We wanted the example texts to be as representative of the whole dataset as possible. Then, for the one-shot scenario, we calculated the distance of each text from the midpoint on their corresponding emotional scales for each emotion separately. By combining these two types of metrics, we have then picked the texts that are both the most representative in terms of vector similarity and were rated to express their corresponding emotional constructs in neither a high nor low manner. We repeated the same operation for the two-shot scenario. However, the texts were picked based on the distance from the lowest point (first text) and the highest point (second text) on the emotional scale. The three-shot scenario combined the examples from one-shot and two-shot. The four-shot scenario picked texts were picked based on distance from the 0.20, 0.40, 0.60, and 0.80 points of the emotional scale. Finally, the five-shot scenario texts were picked based on the distance from the distance from the points of the emotional scale represented as the following fraction points: 1/6, 2/6, 3/6, 4/6, 5/6. The logic was to gradually present the LLM with a more fine-grained representation of the emotional spectrum.

There were multiple cases where the LLM did not respond to the request with an intelligible number, either refusing to honor the request based on the query not complying with OPENAI regulations or simply saying that it cannot assess the emotionality of the specific snippet. We considered this when picking the best multiple-shot scenario for each emotion taxonomy. The test results in the basic emotions condition showed that the three-shot method reared the best results for this task (see Table 1). The averaged correlation between the actual data and the scores provided by the LLM for all basic emotions achieved the highest level for this setting (r = 0.72). The averaged standard deviation of the scores for all basic emotions for this setting was lower than zero-shot and higher than two-shot (zero-shot: SD = 1.53; one-shot: SD = 1.15; two-shot: SD = 1.10). However, we chose to focus on correlation as the decisive metric. The total rejected texts for this scenario were also considerably low, totaling only 47 texts across all emotions.

\begin{table}[h!]
  \centering
  \caption{Zero to five-shot method in basic emotion conditions - Pearson’s r correlations and descriptive statistics}
  \label{tab:emotion_stats}
      \renewcommand{\arraystretch}{1.5}
  \begin{tabular}{lccc}
    \hline
    Type & \( r \) & SD & \( n \) Rejected \\ 
    \hline
    Zero Shot & 0.62 & 1.53 & 0 \\
    One Shot & 0.66 & 1.15 & 339 \\
    Two Shot & 0.66 & 1.10 & 16 \\
    Three Shot & 0.72 & 1.19 & 47 \\
    Four Shot & 0.68 & 1.06 & 57 \\
    Five Shot & 0.68 & 1.15 & 230 \\
    \hline
  \end{tabular}
\end{table}

The dimension-oriented tests pointed towards the two-shot scenario as most applicable for their setting (see Table 2). Here, the two-shot scenario had the highest averaged correlation (r = 0.77) while, at the same time, an acceptable averaged standard deviation of scores (SD = 1.28). The total rejected texts for this scenario were also considerably low totaling 27 texts.

\begin{table}[h!]
  \centering
  \caption{Zero to five-shot method in dimension-oriented conditions - Pearson’s r correlations and descriptive statistics}
  \label{tab:dimension_stats}
      \renewcommand{\arraystretch}{1.5}
  \begin{tabular}{lccc}
    \hline
    Type & \( r \) & SD & \( n \) Rejected \\ 
    \hline
    Zero Shot & 0.72 & 1.20 & 3 \\
    One Shot & 0.71 & 1.61 & 59 \\
    Two Shot & 0.77 & 1.28 & 27 \\
    Three Shot & 0.73 & 1.37 & 140 \\
    Four Shot & 0.72 & 1.24 & 87 \\
    Five Shot & 0.74 & 1.26 & 483 \\
    \hline
  \end{tabular}
\end{table}

Concluding, the three-shot scenario was chosen for the basic emotion setup, while for the dimensional taxonomy setup the two-shot we have picked the two-shot method. These methods were then used to annotate the test set using both GPT3.5 and GPT4.

\subsubsection{Costs}
The participants in the annotation process were paid around \$2,400 in total, split equally between them. At the same time, the calls to the API that were required to perform the multiple shot search totaled \$8.38. The test set annotations, on the other hand, cost us \$65.6, which was driven mostly by the GPT4 API calls.

\section{Results}
The ICC results were presented in Table 3. The reliability of individual rater's assessments ranged from poor to moderate across the tested emotions, with ICC (1) values extending from 0.29 for arousal to 0.60 for valence. In contrast, the reliability of average ratings from multiple raters indicated moderate to good consistency, with ICC (1, k) values ranging from 0.63 for fear to 0.88 for valence. 

\begin{table}[h!]
  \centering
  \caption{The results of the agreement check using the Intraclass Correlation (ICC) statistic}
  \label{tab:icc_results}
    \renewcommand{\arraystretch}{1.5}
  \begin{tabular}{llccc}
    \hline
    Emotion & Type & ICC & \multicolumn{2}{c}{95\% CI} \\ 
    \hline
    Happiness & ICC(1) & 0.53 & [0.52, & 0.54] \\
              & ICC(1,k) & 0.85 & [0.85, & 0.85] \\
    Sadness   & ICC(1) & 0.34 & [0.33, & 0.35] \\
              & ICC(1,k) & 0.72 & [0.71, & 0.73] \\
    Anger     & ICC(1) & 0.54 & [0.53, & 0.55] \\
              & ICC(1,k) & 0.85 & [0.85, & 0.86] \\
    Disgust   & ICC(1) & 0.41 & [0.40, & 0.42] \\
              & ICC(1,k) & 0.77 & [0.77, & 0.78] \\
    Fear      & ICC(1) & 0.25 & [0.24, & 0.26] \\
              & ICC(1,k) & 0.63 & [0.61, & 0.64] \\
    Pride     & ICC(1) & 0.39 & [0.38, & 0.40] \\
              & ICC(1,k) & 0.76 & [0.76, & 0.77] \\
    Valence   & ICC(1) & 0.60 & [0.59, & 0.61] \\
              & ICC(1,k) & 0.88 & [0.88, & 0.89] \\
    Arousal   & ICC(1) & 0.29 & [0.28, & 0.30] \\
              & ICC(1,k) & 0.67 & [0.66, & 0.68] \\
    \hline
  \end{tabular}
    \begin{flushleft}
    \textit{Note.} In this table, ICC(1) values assess the reliability of individual raters' assessments, reflecting the degree to which individual measurements are consistent across different raters or sessions. ICC(1,k) values represent the reliability of average ratings from multiple raters, providing a measure of agreement that accounts for the increased reliability typically observed when individual ratings are aggregated.
    \end{flushleft}
\end{table}

\subsection{Supervised model results}
The results of the main model, as summarized in Table 4, demonstrated the model's performance across different emotion categories and two general affect dimensions: valence and arousal. The table presents correlation coefficients and standard deviations (SDs) for the model predictions compared to human annotations, along with the original standard deviations observed in the human annotations.

\begin{table}[h!]
  \centering
  \caption{The results of the main model}
  \label{tab:main_model_results}
  \renewcommand{\arraystretch}{1.5}
  \begin{tabular}{lccc}
    \hline
    Emotion & Correlation & Model’s SD & Annotator’s SD \\
    \hline
    Happiness & 0.87 & 0.22 & 0.26 \\
    Sadness & 0.75 & 0.15 & 0.20 \\
    Anger & 0.85 & 0.24 & 0.30 \\
    Disgust & 0.81 & 0.19 & 0.24 \\
    Fear & 0.73 & 0.11 & 0.17 \\
    Pride & 0.80 & 0.20 & 0.24 \\
    Valence & 0.87 & 0.22 & 0.27 \\
    Arousal & 0.75 & 0.15 & 0.21 \\
    \hline
  \end{tabular}
  \begin{flushleft}
    \textit{Note.} The standard deviations of the annotators were calculated based on the averaged labels for each text.
  \end{flushleft}
\end{table}
The model exhibited strong correlations with human ratings, particularly in predicting happiness and valence, both achieving the highest correlation of 0.87. It indicated a high level of agreement between the model's predictions and human judgments for these emotional dimensions, suggesting that the model was particularly effective at identifying positive emotional content and overall emotional valence.

Correlations for other emotions, such as sadness (r = 0.75), anger (r = 0.85), disgust (r = 0.81), fear (r = 0.73), and pride (r = 0.80), also indicated a substantial agreement with human annotations, although to a slightly lesser extent than happiness and valence. These results suggested that the model can generally capture a wide range of emotional states, with varying degrees of effectiveness across different emotions.

The SDs of the predictions generated by the model (Model’s SD) were consistently lower than those observed in averaged human annotations (“Annotator’s SD”) for all emotions and affect dimensions. This difference in variability indicated that the model's predictions tend to be more consistent than human ratings. For example, the model's predictions for happiness had an SD of 0.22, compared to the original human annotation SD of 0.26. Similar patterns were observed across all emotions and affect dimensions, with the model's predictions showing less variability than averaged human annotations.

In conclusion, the model demonstrated a strong ability to predict human emotional annotations across diverse emotions and affect dimensions. The high correlation values indicated a significant agreement between the model's predictions and human judgments. In contrast, the lower SDs in the model's predictions suggested a higher consistency in the model's performance compared to the variability inherent in human annotations. These results underscored the model's potential in effectively capturing and predicting human emotional responses to textual content.
\subsection{K-fold validation}
The results of the k-fold validation are presented in Table 5.

\begin{table}[h!]
  \centering
  \caption{The results of the 10 k-fold robustness check}
  \label{tab:k_fold_robustness}
  \renewcommand{\arraystretch}{1.5}
  \makebox[\textwidth][c]{
  \begin{tabular}{lcccccccc}
    \hline
    Emotion & Happiness & Sadness & Anger & Disgust & Fear & Pride & Valence & Arousal \\
    \hline
    Mean & 0.83 & 0.68 & 0.81 & 0.75 & 0.67 & 0.76 & 0.84 & 0.71 \\
    Correlation & [0.82, 0.84] & [0.68, 0.70] & [0.80, 0.82] & [0.74, 0.77] & [0.65, 0.69] & [0.74, 0.77] & [0.83, 0.85] & [0.70, 0.72] \\
    and CI 95\% &  & &  &  & & &  & \\
    \hline
  \end{tabular}}
\end{table}

In assessing the robustness of the supervised model, we conducted a 10-fold cross-validation, focusing on a spectrum of emotional dimensions and affective states, including happiness, sadness, anger, disgust, fear, pride, valence, and arousal. The results revealed a generally high level of reliability across these dimensions. Specifically, the model exhibited strong performance in identifying happiness (mean correlation of 0.83, for 95\% CI see Table 5), anger (r = 0.81), and valence (r = 0.84), indicating a consistent ability to assess these emotional states across different data subsets.

Moderate to strong correlations were observed for sadness (r = 0.68), disgust (r = 0.75), fear (r = 0.67), pride (r = 0.76), and arousal (r = 0.71), with the confidence intervals suggesting a stable performance across the folds, albeit with some variability, particularly in detecting fear. These outcomes did not highlight the supervised model's overall reliability and generalizability.

\subsection{LLM Annotation Results}
The LLM annotation attempts explored two distinct scenarios: a two-shot setup for assessing valence and arousal, and a three-shot setup for discrete emotions including happiness, sadness, anger, fear, disgust, and pride. The two-shot approach involved GPT3.5 two-shot and a variant leveraging GPT4, while the three-shot scenario explored a GPT3.5 three-shot alongside a three-shot GPT4 variant. These setups were selected based on prior tests to optimize the LLM's performance in emotion annotation.

In the two-shot setup for valence and arousal (Table 6), the GPT3.5 two-shot approach yielded correlations of 0.79 for valence and 0.53 for arousal, with standard deviations (SD) of 1.30 and 1.02, respectively. The two-shot GPT4 variant showed slightly improved performance with correlations of 0.79 for valence and 0.55 for arousal, and reduced SDs of 1.23 and 0.93, respectively. Notably, the GPT4 variant exhibited no rejected texts, suggesting enhanced reliability or acceptance criteria compared to the GPT3.5 two-shot setup, which had 36 rejected texts for valence and 2 for arousal.

The three-shot scenario focused on discrete emotions (Table 7) demonstrated varied performance across different emotions. The GPT3.5 three-shot approach showed correlations ranging from 0.46 for fear (the lowest) to 0.78 for anger (the highest), with corresponding SDs spanning from 1.22 for pride to 1.73 for fear. The number of rejected texts varied significantly across emotions, with fear seeing the highest rejection at 12 texts, indicating potential challenges in consistently annotating this emotion.

The three-shot GPT4 variant, however, marked a noticeable improvement in both correlation and SD across all emotions, with correlations improving to 0.88 for happiness and 0.83 for anger, among others. SDs were generally lower, indicating more consistent annotations, with pride having the lowest SD at 0.86. Remarkably, this variant showed no rejected texts across all emotions, underscoring its enhanced capability in emotion annotation.

\begin{table}[h!]
  \centering
  \caption{Dimensions of valence and arousal - results of LLM annotation (Pearson’s correlations and standard deviations)}
  \label{tab:llm_annotation_results}
  \renewcommand{\arraystretch}{1.5}  
  \begin{tabular}{lcc}
    \hline
    Type & Valence & Arousal \\
    \hline
    Two shot GPT3.5 & 0.79 (1.29) & 0.53 (1.02) \\
    Two shot GPT4 & 0.78 (1.23) & 0.55 (0.93) \\
    \hline
  \end{tabular}

  \begin{flushleft}
  \textit{Note.} For the two-shot method, 36 and 2 texts were rejected, respectively, for valence and arousal. For the two-shot GPT4 method, none of the texts was excluded either for valence or arousal. Standard deviations were reported in parentheses.
  \end{flushleft}
\end{table}

\begin{table}[h!]
  \centering
  \caption{Discrete emotions - results of LLM annotation (Pearson’s correlations and standard deviations)}
  \label{tab:discrete_emotions}
  \renewcommand{\arraystretch}{1.5}  
  \begin{tabular}{lcccccc}
    \hline
    Type & Happiness & Sadness & Anger & Fear & Disgust & Pride \\
    \hline
    Two shot GPT3.5 & 0.76 & 0.59 & 0.78 & 0.46 & 0.70 & 0.49 \\
                    & (1.40) & (1.42) & (1.65) & (1.72) & (1.65) & (1.22) \\
    Two shot GPT4   & 0.88 & 0.66 & 0.83 & 0.65 & 0.72 & 0.67 \\
                    & (1.12) & (1.00) & (1.21) & (1.09) & (1.00) & (0.85) \\
    \hline
  \end{tabular}

  \begin{flushleft}
  \textit{Note.} For three-shot method, numbers of rejected texts were as follows: happiness: n = 3, sadness: n = 2, anger: n = 5, fear: n = 12, disgust: n = 1, and pride: n = 16. For three-shot GPT4 method, none of the texts was excluded for all emotions. Standard deviations were reported in parentheses.
  \end{flushleft}
\end{table}

As can be seen in Figure 1. the distribution of labels varies between the original annotators, the GPT-3.5 generated labels, and those generated by GPT-4 varies considerably. Starting with the annotators, their labels for basic emotions were highly skewed towards lower values, with the label of 1 being the most popular. In contrast, the distributions for valence and arousal approached a monotonic shape, with the exception of the two most extreme labels 4, and 5, which were less numerous. In comparison, GPT3.5 exhibited a bimodal distribution for basic emotions, an uneven distribution for valence where “4” was the least used label, and a centered at the middle, mostly leptokurtic distribution for arousal with a sudden increase in counts on the “5” label. GPT3.5’s distributions were therefore visibly different than the original annotators’ distributions. On the other hand, GPT4’s distributions had a significantly better alignment with the original ones, being similarly skewed towards the lower values for basic emotions, without a pronounced bimodal peak at both ends of the spectrum. For valence, GPT-4’s distribution was bimodal with leptokurtic characteristics different from both GPT3.5’s and original annotator’s distributions. Finally, for arousal GPT4’s distribution overlapped with the original annotator’s distribution pretty well, apart from a slight peak at the value of “2” and a drop at the value of “4”.

\begin{figure}[h!] 
  \textbf{Figure 1. Histograms of annotators’, GPT3.5’s, and GPT4’s labels for specific emotion metrics} 
  \centering 
  \includegraphics[width=1\textwidth]{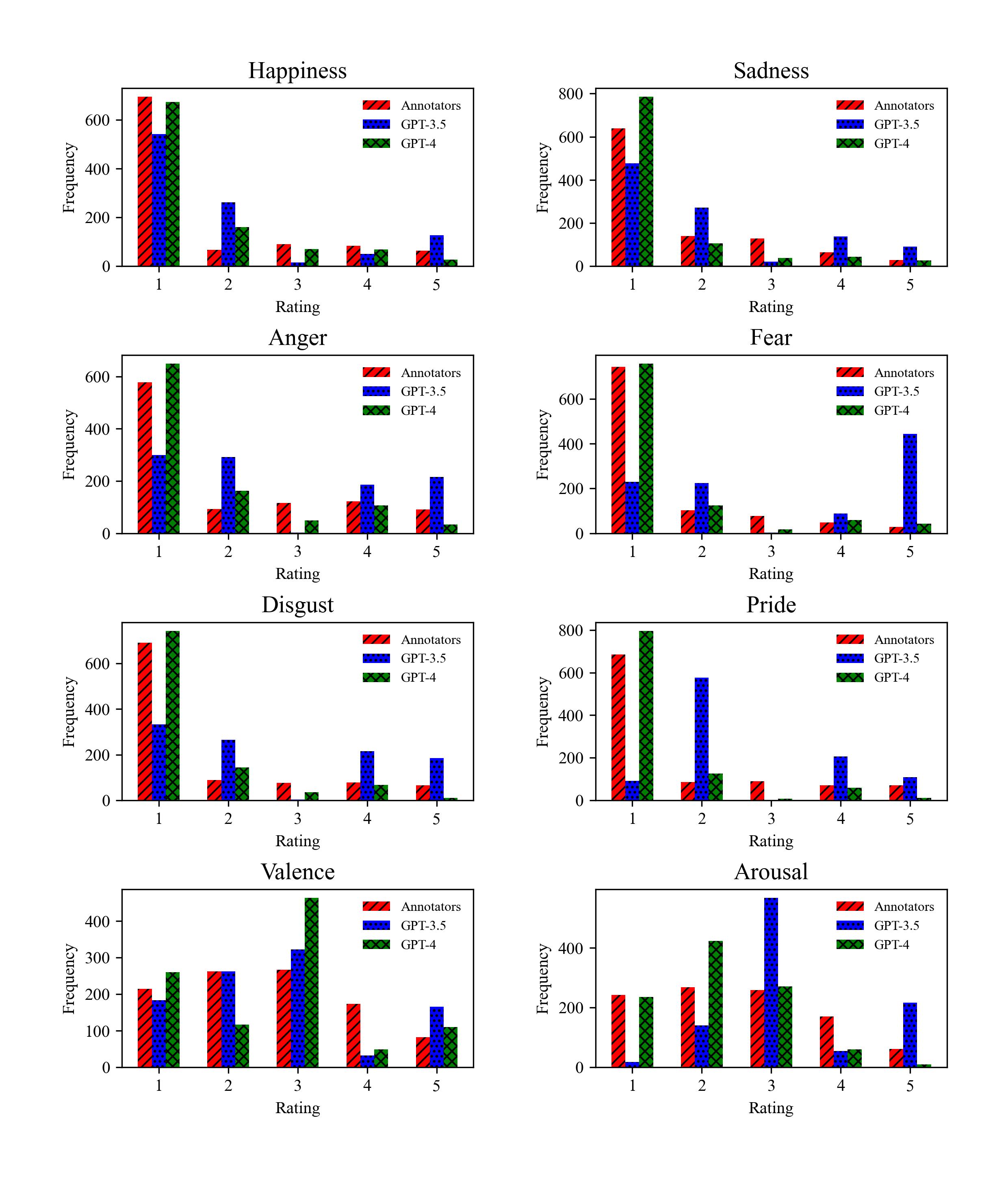} 
  \label{fig:fig1} 
  \caption{The figure displays histograms created for three types of annotations: those made by the original raters, and those created by both GPT-3.5 and GPT-4. In order to compare these distributions directly, the original annotator labels, before averaging, were used to create the histogram. As each text was labeled by exactly 5 annotators, these labels were scaled by dividing by 5 to make them comparable to the labels generated by the LLMs.}
\end{figure}

\newpage
\subsection{Direct comparison}
For the direct comparison we took the best performing LLM model results for both emotion categories, which was the GPT4. As can be seen in Table 8, for happiness, the GPT4 variant slightly outperformed the supervised model with a correlation of 0.88 (SD = 1.12) compared to the supervised model's 0.87 (SD = 0.22). This indicated a marginally higher accuracy in the GPT4 model, albeit with increased variability.

In the case of sadness, the supervised model exhibited a higher correlation of 0.75 (SD = 0.15) relative to the GPT4 variant's 0.66 (SD = 1.00), suggesting the supervised model's superior ability to accurately annotate sadness with less variability. For anger, the supervised model also showed a higher correlation of 0.85 (SD = 0.24) against the GPT4 variant's 0.83 (SD = 1.21), indicating a slight edge in accurately capturing expressions of anger, despite the GPT4 variant's broader range of responses. When assessing fear, the supervised model demonstrated a significantly higher correlation of 0.81 (SD = 0.19) compared to the GPT4 variant's 0.65 (SD = 1.09), underscoring the supervised model's enhanced capability in identifying fear-related expressions with greater consistency. For disgust, the correlation values were more similar, with the supervised model at 0.73 (SD = 0.11) and the GPT4 variant at 0.72 (SD = 1.00), suggesting comparable performance levels, though the GPT4 model exhibits greater variability. In evaluating pride, the supervised model's correlation of 0.80 (SD = 0.20) surpassed the GPT4 variant's 0.67 (SD = 0.85), indicating the supervised model's better performance in consistently capturing expressions of pride.

Regarding valence, both models showed equivalent top performance with a correlation of 0.87 for the supervised model (SD = 0.22) and 0.88 for the GPT4 variant (SD = 1.12), albeit with the GPT4 variant displaying higher variability. For arousal, the supervised model's correlation of 0.75 (SD = 0.15) was notably higher than the GPT4 variant's 0.66 (SD = 1.00), indicating the supervised model's superior accuracy and consistency in annotating arousal.

In summary, while the GPT4 variant demonstrated competitive or slightly superior performance in some respects (particularly for happiness and valence), the supervised model generally exhibited higher accuracy and significantly lower variability across most emotions and affective states, highlighting its robustness and reliability in emotion annotation tasks. The standard deviations of GPT4’s predictions, on the other hand, were more similar to the standard deviations of original annotations, before they were averaged to produce training data, while the standard deviations of the supervised model, mirrored those of the averaged labels on which it was trained.

\begin{table}[h!]
  \centering
  \caption{Direct comparison of GPT4 vs. supervised model - Pearson’s correlations and standard deviations}
  \label{tab:comparison_gpt4_supervised}
  \renewcommand{\arraystretch}{1.5}  
  \begin{tabular}{lcccccccc}
    \hline
    Type & Happiness & Sadness & Anger & Fear & Disgust & Pride & Valence & Arousal \\
    \hline
    Supervised & 0.87 & 0.75 & 0.85 & 0.81 & 0.73 & 0.80 & 0.87 & 0.75 \\
         &(0.22) &(0.15) &(0.24) &(0.19) &(0.11) &(0.20) & (0.22) &(0.15) \\
    GPT4 & 0.88 & 0.66& 0.83& 0.65 & 0.72 & 0.67 & 0.88 & 0.66 \\
        & (1.12) & (1.00) & (1.21) & (1.09) & (1.00) & (0.85) & (1.12) & (1.00) \\
    Annotator’s SD \\ after averaging & 0.26 & 0.20 & 0.30 & 0.24 & 0.17 & 0.24 & 0.27 & 0.21 \\
    Annotator’s SD \\ before averaging & 1.28 & 1.10 & 1.4 & 1.00 & 1.27 & 1.27 & 1.23 & 1.20 \\
    \hline
  \end{tabular}
  \begin{flushleft}
  \textit{Note.} Standard deviations are reported in parentheses. Standard deviations for the original annotations were reported for both averaged, and raw labels.
  \end{flushleft}
\end{table}

\section{Discussion}
As the results indicated, the question of whether researchers should use existing LLM models when annotating political texts in low-resource languages such as Polish is nuanced. On the one hand, the supervised models provided marginally, yet visibly, more accurate results. They were either just as good as GPT-4 (in cases of happiness, disgust, and arousal) or better (for all other emotions). While the standard deviation of the LLMs' predictions was more similar to individual, non-aggregated labels, it is not clear whether this should be considered an asset. This is because the standard deviations of arguably more representative, aggregated human emotionality labels were far smaller. These smaller values were mirrored in the distribution of the predictive model’s labels. One significant advantage of the supervised model is its resilience to external circumstances, such as API availability. Once trained, the model can be stored on the researchers' machines and reused at any time for practically free (excluding computing costs). The availability of the API, while not completely uncertain, is less reliable.

On the other hand, the supervised models require a costly annotation process that is orders of magnitude more resource intensive. One significant upside of this annotation process is that the data gathered can be opened to the large public as we do so for this paper, and thus reused for different projects. The annotation issue is further complicated by the fact that without gathering at least some annotations it is hard to estimate the reliability of LLMs for the specific task that it is supposed to be used for. Therefore, it is hard to avoid this laborious process. However, for evaluating the performance of the LLM without previous parameter searches with regards to a specific multiple shot setup the size of the annotated dataset can be significantly smaller than that required for supervised learning and can perhaps be carried out by the researchers themselves. Of course, a smaller dataset also makes it difficult to choose examples for the multiple shot setup. It also limits the possibility of the parameter search for the prompting technique which, when not carried out on a separate validation set, can result in overfitting.

These considerations imply that the preference for use of either approach largely depends on the availability of resources, both of financial and substantive nature. Research programs as well as commercial projects that have the option to engage in large scale annotation projects and train their own models will be rewarded for doing so by higher accuracy of their predictions, as well as more confidence in the long-term utility and reliability of their predictive solutions. On the other hand, those teams which either do not have the resources necessary or do not want to spend them can opt for the LLM-based approach, which will be marginally worse in performance but at the same time offers a fairly easier and faster-to-implement solution. 

The nature of the task such research teams strive to accomplish can thus be considered as another guide to choosing which approach works best for the team. Tasks that permit forgoing some accuracy for the sake of fast resolution are therefore best suited for the LLM approach, while those in which small accuracy errors can propagate and multiply should be tackled with the supervised method. Another important issue is the amount of data that has to be assessed. LLM approaches, while simpler and faster to implement, can run into scaling issues. This has to be considered before choosing the approach by estimating the number of predictions that need to be made for the project and checking the current prices of OPENAI calls. While for research purposes the cost of API calls will rarely be higher than the cost of the annotation process, this might be of greater import to commercial projects. The study's findings need to be viewed in light of the fast pace at which Large Language Models (LLMs) are evolving, which could affect the relevance of our results over time. As new models are developed, the performance and capabilities of LLMs might change, potentially limiting the applicability of our current conclusions. However, by making our code publicly available, we allow for the replication and updating of this study by others, which helps in maintaining the relevance of the findings despite the rapid advancements in the field. The supervised model training code is available at \url{https://colab.research.google.com/drive/1ZIMIicDyEUVA-kHNXfH0oiUAPIVXCZyh?usp=drive_link} while the rest of the code, including LLM querying can be found at \url{https://github.com/hplisiecki/Predicting-Emotion-Intensity-in-Polish-Political-Texts}. We also welcome other researchers to use the pretrained model introduced in this paper. To let them do that we have published it under the following url \url{https://huggingface.co/hplisiecki/polemo_intensity}. Since the data used to train and validate the model come from social media profiles we choose to not publish it at this stage due to legal concerns, although we are working on making it available in the future.

Future research could explore whether the findings of this study also hold for other resource-poor languages and, potentially, other continuous features. Also, one could add another approach to the comparison, involving machine translation into a language with existing labeled data, like English, to see if that is a viable option at least for some problems (Licht et al., 2024).             

\section{Funding}
This research is funded by a grant from the National Science Centre (NCN) 'Research Laboratory for Digital Social Sciences' (SONATA BIS-10, No. UMO-020/38/E/HS6/00302).

\bibliographystyle{unsrt}  

\newpage
\appendix
\section{Appendix}
\subsection{LLM Prompts}
The prompts used for the LLM annotation process were structured as follows:

\textbf{Basic emotions (Happiness, Sadness, Anger, Disgust, Fear, Pride): }

Translation:

\textit{"To what extent does the text below manifest the emotion '{emotion}'? Respond using a 5-point scale, where 1 means the emotion is not present at all and 5 means the emotion is very distinctly present. Please respond with a single number. Text: '{text}' Your response:"}

Original:

\textit{“Na ile przedstawiony poniżej tekst manifestuje emocje "{emotion}". Odpowiedz używając 5 stopniowej skali, gdzie 1 - emocja wogóle nie występuje a 5 - emocja jest bardzo wyraźnie obecna. Odpowiadaj za pomocą pojedynczego numeru. Tekst: "{text}" Twoja odpowiedź:”}

\textbf{Valence}

Translation:

\textit{"What emotional valence do you read in the following text? Respond using a 5-point scale, where 1 indicates a negative emotion is present and 5 indicates a positive emotion is present. Please respond with a single number."}

Original

\textit{“Jaki znak emocji wyczytujesz w poniższym tekście? Odpowiedz używając 5 stopniowej skali, gdzie 1 - obecna jest negatywna emocja a 5 - obecna jest pozytywna emocja. Odpowiadaj za pomocą pojedynczego numeru.”}

\textbf{Arousal}

Translation:

\textit{"What level of arousal do you read in the following text? Respond using a 5-point scale, where 1 means no arousal and 5 means extreme arousal. Please respond with a single number."}

Original:

\textit{“Jaki poziom pobudzenia wyczytujesz w poniższym tekście? Odpowiedz używając 5 stopniowej skali, gdzie 1 - brak pobudzenia a 5 - ekstremalne pobudzenie. Odpowiadaj za pomocą pojedynczego numeru.“}

For the multiple-shot scenarios, the exemplars were added by appending them to the end of the queries above. They had the following structure: 

Translation:

\textit{"Text \{text number\}: """\{text\}""" Your response: """\{score\}""" \#\#\#"}

Original:

\textit{“Tekst \{text number\}: """\{text\}""" Twoja odpowiedź: """\{score\}""" \#\#\#”}

The target text was finally appended in the same manner as the exemplars:

Translation:

\textit{"Text \{text number\}: """\{text\}""" Your response: "}

Original:

\textit{“Tekst \{text number\}: """\{text\}""" Twoja odpowiedź: ”}

\subsection{Annotation Process - instruction for annotators}

Translation:

You will evaluate the emotional content displayed in some short texts.

Your task will be to mark on a five-point scale the degree to which you think that a given sentence is characterized by each of the following emotions: joy, sadness, anger, disgust, fear, and pride. Use a five-point scale as described below:

0 - the emotion does not occur at all

1 - low level of emotion

2 - moderate level of emotion

3 - high level of emotion

4 - very high level of emotion.

Then, we will ask you to estimate the intensity of two additional emotion parameters: the direction of sensations (negative versus positive) and emotional arousal (no arousal versus extreme arousal). On the next screen you will learn the definitions of both parameters and how you will evaluate them.

Read the descriptions of two emotion parameters: the sign of sensations and emotional arousal. You can do this several times to make sure you understand them - it will make it easier for you to complete the task ahead of you.

You will rate each of the emotion dimensions described above on a five-point scale. To make it easier to imagine the states we have in mind, you can use pictograms symbolizing different directions of experiences and the intensity of the emotional states.

For the direction of sensations, use the following scale:
\begin{figure}[h!] 
  \centering 
  \includegraphics[width=0.8\textwidth]{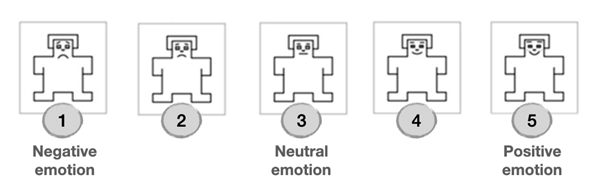} 
  \label{fig:fig2} 
\end{figure}
The first pictogram shows a person who is visibly depressed - specific experiences may include: panic, irritation, disgust, despair, failure, or crisis. The last image shows a person who is visibly excited - specific experiences may include: fun, delight, happiness, relaxation, satisfaction, or rest. The remaining pictograms represent intermediate states.

For emotional arousal, use the following scale:
\begin{figure}[h!] 
  \centering 
  \includegraphics[width=0.8\textwidth]{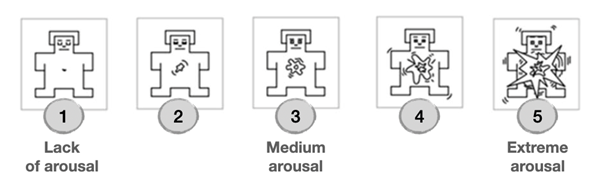} 
  \label{fig:fig3} 
\end{figure}
The first pictogram shows a person who is very calm, almost sleepy - specific experiences may include: relaxation, calm, inactivity, meditation, boredom, or laziness. The last image shows a person who is intensely aroused - appropriate emotional states may include: excitement, euphoria, arousal, rage, agitation, or anger.

Save the link to this manual for later - you can return to it at any time during the examination.

Very important: you can take a break while assessing your statements and return to them at any time - your current work will be saved and you will be able to resume it after the break. If you want to do this, in the upper right corner of the screen you will find the option: "Postpone for later" - click on it, enter the data necessary to save, and confirm the operation. In case you are ready to get back to work: when you enter the study page, an option "Load unfinished survey" will appear in the upper right corner of the screen - select it to load your work.

\subsection{Social Media profiles}
In this research we have scraped the posts of following:

A) Journalists:

Adrian Klarenbach, Agnieszka Gozdyra, Bartosz T. Wieliński, Bartosz Węglarczyk, Bianka Mikołajewska, Cezary Krysztopa, Daniel Liszkiewicz, Dawid Wildstein, Dominika Długosz, Dominika Wielowieyska, Ewa Siedlecka, Jacek Karnowski, Jacek Kurski, Jacek Nizinkiewicz, Janusz Schwertner, Jarosław Olechowski, Konrad Piasecki, Krzysztof Ziemiec, Łukasz Bok, Łukasz Warzecha, Magdalena Ogórek, Magdalena Rigamonti, Marcin Gutowski, Marcin Wolski, Michał Karnowski, Michał Kolanko, Michał Rachoń, Miłosz Kłeczek, Paweł Żuchowski, Piotr Kraśko, Piotr Semka, Radomir Wit, Rafał Ziemkiewicz, Renata Grochal, Robert Mazurek, Samuel Pereira, Szymon Jadczak, Tomasz Lis, Tomasz Sakiewicz, Tomasz Sekielski, Tomasz Sommer, Tomasz Terlikowski, Wojciech Bojanowski, Agaton Koziński, Piotr Witwicki, Jacek Tacik, Magdalena Lucyan, Agata Adamek, Kamil Dziubka, Jarosław Kurski, Dorota Kania, Ewa Bugala, Zuzanna Dąbrowska, Karol Gac, Marcin Tulicki, Marzena Nykiel, Jacek Prusinowski, Paweł Wroński

B) Politicians:

Donald Tusk, Andrzej Duda, Rafał Trzaskowski, Mateusz Morawiecki, Sławomir Mentzen, Janusz Korwin-Mikke, Grzegorz Braun, Szymon Hołownia, Radosław Sikorski, Krzysztof Bosak, Władysław Kosiniak-Kamysz, Borys Budka, Artur E. Dziambor, Marek Belka, Leszek Miller, Mariusz Błaszczak, Roman Giertych, Franek Sterczewski, Konrad Berkowicz, Marek Jakubiak, Michał Szczerba, Przemysław Czarnek, Zbigniew Ziobro, Krzysztof Brejza, Leszek Balcerowicz, Izabela Leszczyna, Klaudia Jachira, Janusz Piechociński, Patryk Jaki, Robert Biedroń, Krystyna Pawłowicz, Katarzyna Lubnauer, Anna Maria Sierakowska, Łukasz Kohut, Marcin Kierwiński, Anna Maria Żukowska, Marian Banaś, Dariusz Joński, Kamila Gasiuk-Pihowicz, Barbara Nowacka, Adrian Zandberg, Krzysztof Śmieszek, Paulina Matysiak, Paweł Kukiz, Michał Wójcik, Sebastian Kaleta, Małgorzata Wassermann, Joachim Brudziński, Maciej Konieczny, Marcelina Zawisza

C) NGOs:

Polska Akcja Humanitarna, Helsińska Fundacja Praw Człowieka, Polski Czerwony Krzyż, Fundacja Dialog, Fundacja Ocalenie, Fundacja Ogólnopolski Strajk Kobiet, Stowarzyszenie Amnesty International, Fundacja Centrum Praw Kobiet, Stowarzyszenie Sędziów Polskich IUSTITIA, Stowarzyszenie Marsz Niepodległości, Lekarze bez Granic, Fundacja TVN, Fundacja Dzieciom "Zdążyć z Pomocą", Wielka Orkiestra Świątecznej Pomocy, Szlachetna Paczka, Fundacja WWF Polska, Fundacja Greenpeace Polska, Liga Ochrony Przyrody, Związek Stowarzyszeń Polska Zielona Sieć, Młodzieżowy Strajk Klimatyczny, Stowarzyszenie Miłość Nie Wyklucza, Kampania Przeciw Homofobii, Stowarzyszenie Lambda - Warszawa, Fundacja Trans-Fuzja, Stowarzyszenie Grupa Stonewall.

\end{document}